\documentclass[sigconf]{acmart}

\usepackage{subcaption}
\newtheorem{definition}{Definition}

\AtBeginDocument{%
  }

\setcopyright{acmlicensed}
\copyrightyear{2018}
\acmYear{2018}
\acmDOI{XXXXXXX.XXXXXXX}
\acmConference[Conference acronym 'XX]{Make sure to enter the correct
  conference title from your rights confirmation email}{June 03--05,
  2018}{Woodstock, NY}
\acmISBN{978-1-4503-XXXX-X/2018/06}

\begin{document}

\title{Kernel Representation and Similarity Measure for Incomplete Data}

\author{Yang Cao}
\affiliation{%
  \institution{Tsinghua University, \\ Great Bay University}
  \state{Guangdong}
  \country{China}
}
\email{charles.cao@ieee.org}

\author{Sikun Yang}
\affiliation{%
  \institution{Great Bay University}
  \city{Guangdong}
  \country{China}}

\author{Kai He}
\affiliation{%
 \institution{Dongguan University of Technology	}
 \state{Guangdong}
 \country{China}}

\author{Wenjun Ma}
\affiliation{%
  \institution{South China Normal University	}
  % \city{Haidian Qu}
  \state{Guangdong}
  \country{China}}

\author{Ming Liu}
\affiliation{%
  \institution{Deakin University}
  % \city{San Antonio}
  \state{VIC}
  \country{Australia}}

\author{Yujiu Yang}
\affiliation{%
  \institution{Tsinghua University}
  \city{Guangdong}
  \country{China}}

\author{Jian Weng}
\affiliation{%
  \institution{Jinan University	}
  \city{Guangdong}
  \country{China}}

\renewcommand{\shortauthors}{Trovato et al.}

\begin{abstract}
% Measuring similarity between incomplete data is a fundamental challenge in web mining, recommendation systems, and user behavior analysis. Traditional approaches either discard incomplete data or perform imputation as a preprocessing step, leading to information loss and biased similarity estimates. This paper presents the proximity kernel that measures similarity between incomplete data in a kernel feature space. The method introduces data-dependent binning and proximity assignment to project data into high-dimensional space, and employs a cascading fallback strategy to estimate missing data distributions in the kernel space. We conduct clustering tasks on the proposed kernel representation across 12 real world incomplete datasets, demonstrating superior performance compared to existing methods while maintaining linear time complexity.

Measuring similarity between incomplete data is a fundamental challenge in web mining, recommendation systems, and user behavior analysis. Traditional approaches either discard incomplete data or perform imputation as a preprocessing step, leading to information loss and biased similarity estimates. This paper presents the proximity kernel, a new similarity measure that directly computes similarity between incomplete data in kernel feature space without explicit imputation in the original space. The proposed method introduces data-dependent binning combined with proximity assignment to project data into a high-dimensional sparse representation that  adapts to local density variations. For missing value handling, we propose a cascading fallback strategy to estimate missing feature distributions. We conduct clustering tasks on the proposed kernel representation across 12 real world incomplete datasets, demonstrating superior performance compared to existing methods while maintaining linear time complexity. All the code are available at \url{https://anonymous.4open.science/r/proximity-kernel-2289}.
\end{abstract}





\maketitle

\section{Introduction}

The proliferation of web applications has led to an unprecedented growth in data collection from diverse sources, including user interactions, sensor networks, and social media platforms. However, real-world data is inherently incomplete due to various factors such as user privacy preferences, survey
non-response and voluntary non-disclosure of information, making the development of robust similarity measures for incomplete data a critical challenge~\cite{altman2007missing}.

Measuring similarity between incomplete data samples is a critical problem in web mining and data analysis. Recommendation systems exemplify this challenge: users typically rate only a small fraction of available items, producing highly sparse user-item matrices~\cite{goksel2025novel}. The problem intensifies with data heterogeneity, as incompleteness can affect numerical, categorical, or mixed-type features simultaneously. This challenge is critical because similarity measurement serves as the foundation for downstream tasks such as clustering, classification and anomaly detection, where inaccurate similarity estimates can significantly degrade task performance~\cite{emmanuel2021survey}.

Traditional approaches to handle missing data in similarity measure fall into several categories. Deletion-based approaches simply ignore missing values or remove incomplete samples entirely. While computationally efficient, these methods suffer from significant information loss and can introduce bias. Imputation-based methods fill missing values using statistical measures (mean, median, mode) or more sophisticated techniques like k-nearest neighbors imputation, matrix factorization, or iterative approaches~\cite{kim2021statistical}. However, these approaches often fail to capture the underlying data distribution and may introduce artificial patterns that do not reflect the true similarity structure.

Recent advances have introduced neural network-based approaches for handling missing data, including autoencoders and generative adversarial networks (GANs) that can learn complex data distributions and generate plausible missing values. While promising, these methods typically require large datasets and substantial computational resources, making them less suitable for many web applications where efficiency and interpretability are crucial. Moreover, they often lack theoretical guarantees about the quality of the learned representations and can be sensitive to hyperparameter choices~\cite{adhikari2022comprehensive}.

In this paper, we introduce a novel similarity measure called proximity kernel specifically designed for incomplete data that addresses the limitations of existing approaches. Our method is built on two key innovations: a data dependent binning mechanism combined with proximity assignment, and a cascading fallback strategy that maximizes the utilization of available information for missing value handling.

The binning strategy combined with proximity assignment exhibits data dependent properties: dense regions result in narrower bins with closer quantiles, while sparse regions produce wider bins with more distant quantiles. Through proximity assignment, observed data points are projected into a high-dimensional sparse space via one-hot encoding. This creates an important characteristic where two points in dense regions may have lower similarity than two points in sparse regions, even when their actual distances are comparable, reflecting the local data structure without explicit density estimation.

For missing value handling, we propose a cascading fallback strategy that operates at three levels in the high-dimensional representation space. Rather than imputing values in the original feature space, we estimate the missing encoding as a probability distribution over bins using kernel mean embedding (KME)~\cite{smola2007hilbert, muandet2017kernel}. First, we seek samples matching across all observed features (intersection-based matching) and use their KME to represent the distribution of missing features. If unsuccessful, we relax to samples matching on any observed feature (union-based matching). Finally, we fall back to global distribution priors. This strategy operates entirely in the representation space, where KME captures the central tendency of distributions without requiring explicit imputation in the original space.

The main contributions of this paper are: 
\begin{itemize}
    \item proposing proximity kernel, a similarity measure that projects data into a high-dimensional sparse representation through equal-frequency binning and Voronoi-based proximity assignment, adapting to local density without explicit density estimation.
    \item introducing a cascading fallback strategy for missing value handling that progressively relaxes matching constraints from intersection to union to global priors.
    \item showing that the proposed method achieves linear time complexity, making it scalable to large-scale datasets.
    \item providing experiments on 12 datasets to demonstrate superior performance over existing methods in clustering tasks.
\end{itemize}

\section{Related Work}

Current approaches to similarity measure for incomplete data primarily rely on imputation methods, which can be broadly categorized into traditional statistical techniques and neural network-based approaches.

\subsection{Traditional Imputation Methods}

Missing data imputation has been extensively studied across various domains. The simplest approach is mean imputation, which replaces missing values with the feature mean for numerical variables or mode for categorical variables~\cite{farhangfar2007novel, miao2022experimental}. While computationally efficient, this method fails to preserve the natural variability of data and can lead to biased estimates when missing data is not random.

K-nearest neighbors (KNN)~\cite{altman1992introduction} imputation estimates missing values based on the average of $k$ most similar complete observations. This approach can capture local patterns but struggles with high-dimensional data and is sensitive to the choice of $k$ and distance metric, often producing poor results when the feature space is sparse.

The Expectation-Maximization (EM)~\cite{ghahramani1993supervised} algorithm provides a principled statistical approach to impute missing values based on maximum likelihood density estimation. However, EM requires strong distributional assumptions that may not hold in practice and can converge to local optima.

Multivariate Imputation by Chained Equations (MICE)~\cite{van2011mice} extends single imputation by creating multiple imputed datasets and combining results to account for imputation uncertainty. MICE can handle mixed-type data and complex missing patterns but requires careful specification of imputation models for each variable and can be computationally expensive.

Tree-based methods like MissForest~\cite{stekhoven2012missforest} employ random forests to iteratively impute missing values, leveraging the ensemble nature of random forests to provide robust estimates. MissForest can handle non-linear relationships and mixed-type data but may overfit when the number of variables is large relative to the sample size and lacks theoretical convergence guarantees.

\subsection{Deep Learning and Generative Approaches}

Recent advances have introduced neural network-based approaches for handling missing data. Generative Adversarial Imputation Networks (GAIN)~\cite{yoon2018gain} employs a minimax game between generator and discriminator networks to learn the distribution of missing data. MIWAE~\cite{mattei2019miwae} handles missing data by using deep latent variable models and maximises a potentially tight lower bound of the log-likelihood of the observed data.  Missing data Imputation Refinement And Causal LEarning (MIRACLE)~\cite{kyono2021miracle} combines causal inference with deep learning to improve imputation quality by leveraging causal relationships among variables. While deep learning-based methods can produce realistic imputations, they require substantial computational resources and large datasets for stable training.

\subsection{Kernel-based Similarity Measure}

Traditional similarity measures such as Euclidean distance are not directly applicable to incomplete data. Pairwise deletion simply ignores missing values when computing distances, but this can lead to biased similarity estimates and significant information loss when missing data is not random.

% Distance-based methods often employ weighted approaches where the contribution of each feature to the overall similarity is adjusted based on data availability. Some methods normalize distances by the number of observed features, while others use more sophisticated weighting schemes based on feature informativeness. However, these approaches often fail to capture the true underlying similarity structure and may be sensitive to the missing data mechanism.

% Probabilistic approaches to similarity measurement model the joint distribution of features and compute similarities based on likelihood ratios or probabilistic distances. These methods can naturally handle missing data by marginalizing over unobserved variables but often require strong distributional assumptions that may not hold in practice and can be computationally prohibitive for high-dimensional data.

HI-PMK~\cite{zhou2025handling} represents a recent advance in kernel methods for incomplete data. This approach employs a data-dependent kernel that measure the similarity between missing data. While HI-PMK provides a principled way to compute similarities for incomplete observations, it suffers from high computational complexity, requires careful hyperparameter tuning, and may not scale well to large datasets.

% A fundamental limitation across all existing approaches is that they treat imputation and similarity computation as separate steps, potentially leading to suboptimal similarity estimates. Methods that impute missing values first and then compute similarities may introduce artifacts that do not reflect true underlying relationships, resulting in similarity measures that are biased toward the imputation model's assumptions rather than the true data structure.

\section{Preliminary}

\subsection{Problem Formulation}

Let $\mathcal{D} = \{\mathbf{x}_1, \mathbf{x}_2, ..., \mathbf{x}_n\}$ be a dataset of $n$ samples, where each sample $\mathbf{x}_i \in \mathbb{R}^d$ is a $d$-dimensional feature vector that may contain missing values. We denote missing values as NaN.

For each sample $i$, let $O_i \subseteq \{1, 2, ..., d\}$ be the set of observed feature indices and $M_i = \{1, 2, ..., d\} \setminus O_i$ be the set of missing feature indices. The observed portion of sample $i$ is $\mathbf{x}_i^{obs} = \{x_{ij} : j \in O_i\}$ and the missing portion is $\mathbf{x}_i^{mis} = \{x_{ij} : j \in M_i\}$.

The goal is to compute a similarity function $s: \mathcal{D} \times \mathcal{D} \rightarrow [0,1]$ that quantifies the similarity between any two samples $\mathbf{x}_i$ and $\mathbf{x}_j$, regardless of their missing values. The resulting similarity can be used for downstream tasks such as clustering, where we partition the dataset into $K$ groups based on the computed similarities.

\subsection{Kernel Methods}

\begin{definition}[Kernel Function]
A kernel function is a symmetric positive definite function $k: \mathcal{X} \times \mathcal{X} \rightarrow \mathbb{R}$ that measures the similarity between two data points in the input space $\mathcal{X}$.
\end{definition}

By Mercer's theorem, any valid kernel function $k(\mathbf{x}, \mathbf{x}')$ can be expressed as an inner product in some reproducing kernel Hilbert space (RKHS) $\mathcal{H}$:
\begin{equation}
    k(\mathbf{x}, \mathbf{x}') = \langle \phi(\mathbf{x}), \phi(\mathbf{x}') \rangle_{\mathcal{H}},
\end{equation}
where $\phi: \mathcal{X} \rightarrow \mathcal{H}$ is a feature map that implicitly maps data points from the input space to a potentially infinite-dimensional feature space. This formulation, known as the kernel trick, enables us to compute inner products in $\mathcal{H}$ without explicitly evaluating the feature map $\phi$, thereby avoiding the computational burden of working directly in high-dimensional spaces. 

\subsection{Kernel Mean Embedding}

\begin{definition}[Kernel Mean Embedding]
Given a probability distribution $P$ over $\mathcal{X}$ and a kernel function $k$ with associated feature map $\phi$, the kernel mean embedding of $P$ is defined as~\cite{smola2007hilbert, muandet2017kernel}:
\begin{equation}
    \mu_P = \mathbb{E}_{\mathbf{x} \sim P}[\phi(\mathbf{x})] \in \mathcal{H}.
\end{equation}
\end{definition}

The kernel mean embedding provides a powerful framework for representing probability distributions as elements in an RKHS. Intuitively, $\mu_P$ can be viewed as the "center" of the distribution $P$ when the data points are mapped into the feature space $\mathcal{H}$ via $\phi$. This representation maps an entire probability distribution to a single point in the high-dimensional feature space, enabling distribution-level operations through simple vector operations in $\mathcal{H}$.

For practical applications with finite sample data $\{\mathbf{x}_i\}_{i=1}^n$ drawn i.i.d. from $P$, the empirical kernel mean embedding is given as:
\begin{equation}
    \hat{\mu}_P = \frac{1}{n}\sum_{i=1}^n \phi(\mathbf{x}_i).
\end{equation}

When the kernel is characteristic (e.g., isolation kernel~\cite{ting2018isolation}), the kernel mean embedding is injective, meaning different distributions are mapped to different points in $\mathcal{H}$, which makes KME a unique and complete representation of distributions.

% \section{Problem Formulation}

% Let $\mathcal{D} = \{x_1, x_2, \ldots, x_n\}$ denote a dataset where each data point $x_i \in \mathbb{R}^d$ represents a $d$-dimensional feature vector that may contain missing values. For each data point $x_i$, let $\mathcal{M}_i \subseteq \{1, 2, \ldots, d\}$ denote the set of observed feature indices, and $\mathcal{U}_i = \{1, 2, \ldots, d\} \setminus \mathcal{M}_i$ represent the missing feature indices. The observed portion is $x_i^{obs} = \{x_{i,j} : j \in \mathcal{M}_i\}$ and the missing portion is $x_i^{mis} = \{x_{i,j} : j \in \mathcal{U}_i\}$.

% The goal is to define a similarity measure $s: \mathbb{R}^d \times \mathbb{R}^d \rightarrow [0, 1]$ that computes meaningful similarity between data points $x_i$ and $x_j$ regardless of their missing patterns. Traditional distance-based measures like $\|x_i - x_j\|_2$ are undefined when missing values are present, while using only commonly observed features $\|x_i^{obs \cap obs} - x_j^{obs \cap obs}\|_2$ may lose critical information.

\section{Methodology}
\begin{table}[htbp]
\centering
\caption{Key Notations}
\label{tab:notations}
\begin{tabular}{ll}
\toprule
\textbf{Notation} & \textbf{Description} \\
\midrule
$X$ & Data matrix of size $n \times d$ \\
$n$ & Number of samples \\
$d$ & Number of features \\
$x_{i,j}$ & Value of feature $j$ for sample $i$ \\
$X_{obs,j}$ & Set of observed (non-missing) values for feature $j$ \\
$n_{bins}$ & Number of bins per feature \\
$c_{j,k}$ & $k$-th bin center for feature $j$ \\
$k$ & Bin index, $k \in \{1, 2, ..., n_{bins}\}$ \\
$b_{i,j}$ & Bin assignment for sample $i$, feature $j$ \\
% $\mathbf{h}_{i,j}$ & One-hot encoding vector for sample $i$, feature $j$ \\
$h_{i,j,k}$ & $k$th element of encoding vector$\mathbf{h}_{i,j}$ \\
$\mathbf{z}_i$ & Complete encoding vector for sample $i$ \\
$M_i$ & Set of missing features for sample $i$ \\
$O_i$ & Set of observed features for sample $i$ \\
$C_{i,j}$ & Samples matching sample $i$ in feature $j$ \\
$S_1(i)$ & Intersection-based matching samples \\
$S_2(i)$ & Union-based matching samples \\
$p_{j,k}$ & Global prior probability for feature $j$, bin $k$ \\
$K(x_i, x_j)$ & Proximity kernel between samples $x_i$ and $x_j$ \\
\bottomrule
\end{tabular}
\end{table}

In this section, we propose proximity kernel that address the challenge of computing meaningful similarities between incomplete data by constructing a data-dependent kernel representation in a high-dimensional space. Given a dataset $X \in \mathbb{R}^{n \times d}$ containing $n$ samples with $d$ features, where missing values are represented as $NaN$, our goal is to construct a kernel mapping that embeds incomplete data into a reproducing kernel Hilbert space (RKHS) where both imputation and similarity measurement can be performed jointly.

The core insight of our approach is that instead of imputing missing values in the original feature space and then computing similarities, we construct a data-dependent kernel that maps samples to a high-dimensional sparse representation where the geometry naturally reflects the underlying data distribution. This kernel representation exhibits a critical density-adaptive property: \emph{two points in a sparse region are more similar than two points of equal inter-point distance in a dense region}. This property emerges from the combination of equal frequency binning and proximity assignment, making the similarity measure adapt to local density without explicit density estimation.

The proposed proximity kernel consists of four main components: equal-frequency binning that selects density-adaptive bin centers, proximity assignment that assigns values to nearest bin centers creating Voronoi-like diagram, one-hot encoding that embeds samples into a sparse high-dimensional space, and a cascading fallback strategy that handles missing values through kernel mean embedding principles at progressively relaxed matching levels. The detailed steps are shown as following:

\subsection{Data Dependent Binning and Proximity Assignment}

\subsubsection{Bin Center Selection via Equal-Frequency Binning}
For each feature $j \in \{1, 2, ..., d\}$, we select $n_{bins}$ bin centers using equal-frequency binning. Let $X_{obs,j}$ denote all observed (non-missing) values for feature $j$. Equal-frequency binning divides the data distribution into $n_{bins}$ intervals, where each interval contains approximately the same number of data points.

The bin centers $c_{j,k}$ are determined by computing evenly-spaced percentiles of the observed distribution:
\begin{equation}
    c_{j,k} = \text{percentile}\left(X_{obs,j}, \frac{(k-1) \times 100}{n_{bins}-1}\right), 
\end{equation}
where $k \in \{1, 2, \dots, n_{bins}\}$ is the index of bins.

This yields $n_{bins}$ centers where $c_{j,1}$ and $c_{j,n_{bins}}$ equal the minimum and maximum values of each feature $j$, and intermediate centers are positioned at equal-frequency intervals. 

The equal-frequency approach ensures that bin centers are positioned where data actually exists, adapting automatically to the underlying distribution. \emph{In dense regions where many data points cluster together, consecutive centers are close in value. In sparse regions with few scattered points, consecutive centers are far apart.} This property forms the foundation for density-adaptive behavior without requiring explicit density estimation.

\subsubsection{Proximity Assignment Mechanism}

Unlike traditional binning methods that assign values to bins based on interval membership, our approach employs proximity assignment. Given the bin centers $\{c_{j,1}, ..., c_{j,n_{bins}}\}$ for each feature $j$, we assign each observed value to its nearest center. For sample $i$ and feature $j$:
\begin{equation}
    b_{i,j} = \arg\min_{k \in \{1,...,n_{bins}\}} |x_{i,j} - c_{j,k}|,
\end{equation}

\noindent where $b_{i,j} \in \{1, ..., n_{bins}\}$ denotes the assigned bin index. This proximity assignment creates a Voronoi diagram of the feature space, where each center $c_{j,k}$ defines a Voronoi cell containing all points closer to it than to any other center.

\begin{figure}
    \centering
    \includegraphics[width=0.95\linewidth]{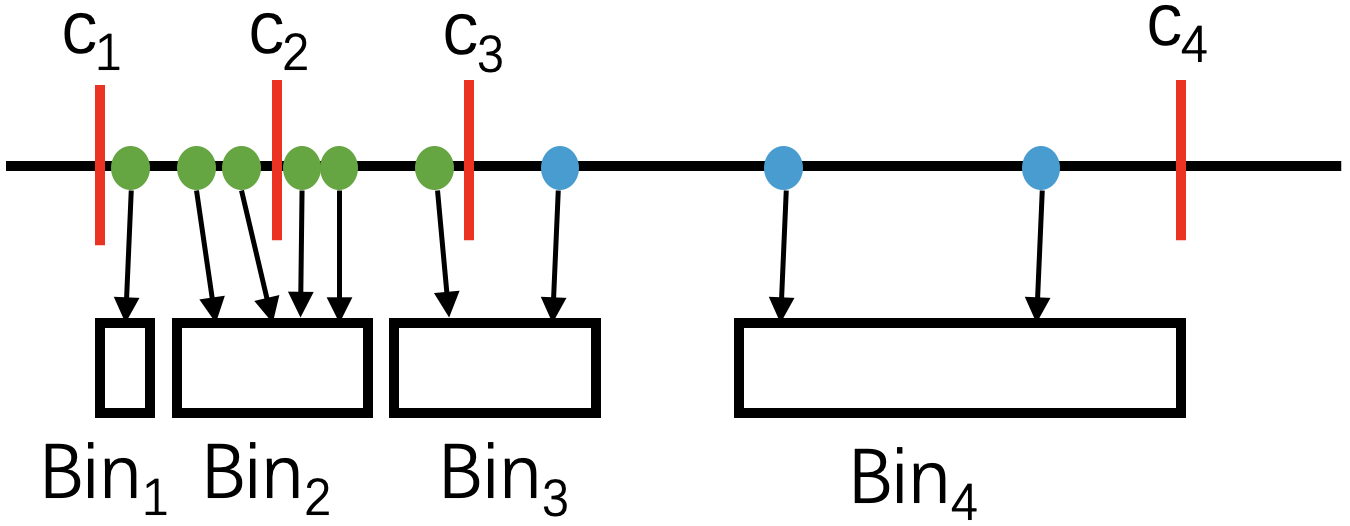}
    \caption{Demonstration of density-adaptive binning. Red lines denote bin centers selected through equal-frequency binning. Green and blue points represent samples in dense and sparse regions, respectively. The sparse region exhibits wider bins compared to the dense region, reflecting the natural adaptation to data distribution without explicit density estimation.}
    \label{fig:binning}
\end{figure}

The combination of equal-frequency centers and proximity assignment produces density-adaptive partitioning. Consider two pairs of points with identical Euclidean distance $\delta$: one pair points in a dense region and another in a sparse region. In the dense region, closely-spaced centers create small Voronoi cells, making it more likely that the two points fall into different cells. In the sparse region, widely-spaced centers create large Voronoi cells, making it more likely that the two points fall into the same cell. 
% Thus, the effective resolution of the discretization adapts to local data density.
Figure~\ref{fig:binning} demonstrates the binning strategy.

\subsubsection{One-Hot Encoding Construction}
After bin assignment, we encode each feature using one-hot representation. For sample $i$ and each feature $j$, we construct a binary vector $\mathbf{h}_{i,j} \in \{0,1\}^{n_{bins}}$:

\begin{equation}
    h_{i,j,k} = \begin{cases} 1 & \text{if } b_{i,j} = k \\ 0 & \text{otherwise} \end{cases}
\end{equation}
% $$h_{i,j} = \begin{cases} 1 & \text{if } b_{i,j} = k \\ 0 & \text{otherwise} \end{cases}$$

For example, if a feature is divided into $n_{bins}=3$ cells. If sample $i$ falls into the first cell for feature $j$, then $\mathbf{h}_{i,j} = [1, 0, 0]$. The complete encoding for sample $i$ concatenates the one-hot vectors across all features as: 
\begin{equation}
    \mathbf{z}_i = [\mathbf{h}_{i,1}, \mathbf{h}_{i,2}, \cdots, \mathbf{h}_{i,d}].    
\end{equation}
Continuing the example with $n_{bins}=3$ and $d=2$ features, if sample $i$ falls into the first cell for feature 1 and the third cell for feature 2, then $\mathbf{z}_i = [1, 0, 0, 0, 0, 1]$.

For samples without missing values, $\mathbf{z}_i$ is sparse with exactly $d$ ones out of $d \times n_{bins}$ total elements, yielding a sparsity ratio of $\frac{n_{bins} - 1}{n_{bins}}$. 
% This sparse representation explicitly encodes which Voronoi cells each feature occupies.

\subsection{Cascading Fallback Strategy for Missing Value}

When sample $i$ contains missing values, we cannot directly apply the proximity assignment and encoding procedures. The core idea is to leverage the observed features to infer which bins the missing features should fall. We identify samples that share similar patterns in their observed features, and use kernel mean embedding over these similar samples to estimate the missing feature distributions.

Let $M_i = \{j : x_{i,j} = \text{NaN}\}$ denote the set of missing features for sample $i$, and $O_i = \{1, ..., d\} \setminus M_i$ denote the observed features. We employ a three-level cascading strategy that progressively relaxes the matching criteria when stricter criteria fail to find similar samples.

\subsubsection{Level 1: Intersection-Based Exact Matching}

The first level seeks samples that match sample $i$ on all observed features. For each observed feature $j \in O_i$, we identify the set $C_{i,j}$ of all other samples (indexed by $m$) that fall into the same bin as sample $i$:
\begin{equation}
    C_{i,j} = \{m : m \neq i, b_{m,j} = b_{i,j}\},
\end{equation}

The intersection-based matching set $S_1(i)$ is the intersection across all observed features:
\begin{equation}
    S_1(i) = \bigcap_{j \in O_i} C_{i,j}.
\end{equation}

This set contains all samples (excluding sample $i$ itself) that have been assigned to the same bins as sample $i$ for each observed feature. If $S_1(i) \neq \emptyset$, we estimate missing encoding through kernel mean embedding:
\begin{equation}
    h_{i,j,k} = \frac{1}{|S_1(i)|} \sum_{m \in S_1(i)} h_{m,j,k}, \quad \forall j \in M_i, k = 1, ..., n_{bins}.
\end{equation}

This averaging process is essentially a kernel mean embedding operation that produces values between 0 and 1, representing the empirical probability distribution over bins for each missing feature based on the patterns observed in similar samples.

\subsubsection{Level 2: Union-Based Partial Matching}

In practice, the intersection $S_1(i)$ may be empty, particularly when the sample exhibits a rare combination of feature values. When no samples match across all observed features simultaneously, we cannot apply the intersection-based estimation. To provide a robust fallback mechanism, we relax the constraint to consider samples that match on any observed feature.

We define the union-based matching set as:

\begin{equation}
    S_2(i) = \bigcup_{j \in O_i} C_{i,j}.
\end{equation}

This set contains all samples that share at least one Voronoi cell with sample $i$ in any observed feature. The union operation aggregates samples with partial similarity rather than requiring complete agreement. If $S_2(i) \neq \emptyset$, we apply kernel mean embedding over this broader set:
\begin{equation}
    h_{i,j,k} = \frac{1}{|S_2(i)|} \sum_{m \in S_2(i)} h_{m,j,k}, \quad \forall j \in M_i, k = 1, ..., n_{bins}.
\end{equation}

The key distinction from Level 1 is the relaxation of matching criteria. Level 1 seeks samples with identical patterns across all observed features, providing high-confidence estimates but potentially finding no matches. Level 2 trades specificity for coverage, utilizing information from any partially similar samples. 
% This averaging over a more diverse set of samples produces estimates that are less conditioned on the specific observed pattern.

\subsubsection{Level 3: Global Prior Distribution}
Even the union-based matching may fail to find any similar samples. This occurs when sample $i$ occupies Voronoi cells that no other samples occupy in any of its observed features, which can happen with outliers, samples in sparse regions, or datasets with high-dimensional sparse patterns. When $S_2(i) = \emptyset$, we cannot leverage local similarity and must use the global distribution of each missing feature.

For each missing feature $j \in M_i$ and each bin $k$, we compute the probability:

\begin{equation}
    p_{j,k} = \frac{\text{count of samples in bin } k \text{ for feature } j}{\text{count of samples with observed feature } j}.
\end{equation}

The missing encoding is set to this probability distribution:

\begin{equation}
    \mathbf{h}_{i,j} = [p_{j,1}, p_{j,2}, ..., p_{j,n_{bins}}],
\end{equation}

\noindent where $\sum_{k=1}^{n_{bins}} p_{j,k} = 1$. For example, if 50\% of samples fall in bin 1, 30\% in bin 2, and 20\% in bin 3 for feature $j$, then $\mathbf{h}_{i,j} = [0.5, 0.3, 0.2]$, representing the probability distribution over bins. 

This strategy simply treats each feature as if it were unrelated to the observed features. The global prior acts as the final safety mechanism, guaranteeing that every sample receives a valid encoding even when no similar samples exist.

\subsection{Kernel Representation and Similarity Measure}
The encoding vectors $\mathbf{z}_i \in [0,1]^{d \times n_{bins}}$ form a sparse kernel representation. For complete data, these are binary; for incomplete data with estimated features, they contain fractional values from the cascading fallback strategy.

We define the proximity kernel between samples $m$ and $n$ as:
\begin{equation}
    K(x_m, x_n) = \frac{1}{d} z_m^T z_n = <z_m, z_n>.
\end{equation}
% $$K(x_m, x_n) = \frac{1}{d} <\mathbf{z}_m \mathbf{z}_n>$$

This similarity measure has a probabilistic interpretation: it computes the expected probability that two samples fall into the same bin, averaged across all features.. For each feature, if both samples occupy the same Voronoi cell (same bin), they contribute 1 to the inner product; otherwise they contribute 0. The normalization by $d$ yields the average co-occurrence rate across all features.

The proximity kernel satisfies $K(x_m, x_n) \in [0, 1]$:
\begin{itemize}
    \item $K(x_m, x_n) = 1$ indicates perfect similarity (co-occurrence in all features).
    \item $K(x_m, x_n) = 0$ indicates complete dissimilarity (no co-occurrence).
    \item Intermediate values represent the fraction of features where samples co-occur.
\end{itemize}

For samples with missing values, the fractional encodings naturally integrate into this framework. The inner product $\mathbf{z}_i \mathbf{z}_j$ computes the sum of element-wise products, where fractional values from kernel mean embedding represent probability distributions over bins.

The sparse kernel representation $\mathbf{z}_i$ serves dual purposes: it enables similarity computation through the proximity kernel, and it can be used directly as input features for downstream machine learning tasks such as clustering, classification, or dimensionality reduction on incomplete data.

\subsection{Kernel Validity Proof}

The proximity kernel $K(x_i, x_j)$ is constructed as an inner product in the representation space. We verify that this construction yields a valid kernel by demonstrating it satisfies Mercer's conditions: symmetry and positive semi-definiteness~\cite{christmann2008support}.

\emph{Proof of Positive Semi-Definiteness:} Consider the gram matrix $\mathbf{K} \in \mathbb{R}^{n \times n}$ with entries $K_{ij} = K(x_i, x_j)$ for samples $x_1, ..., x_n$. Let $\mathbf{Z} \in \mathbb{R}^{n \times (d \times n_{bins})}$ be the matrix where each row is a sample representation $\mathbf{z}_i$. The gram matrix can be expressed as:
\begin{equation}
    \mathbf{K} = \frac{1}{d} \mathbf{Z} \mathbf{Z}^T.    
\end{equation}

The matrix $\mathbf{Z} \mathbf{Z}^T$ is a Gram matrix whose $(i,j)$-th entry equals $\mathbf{z}_i^T \mathbf{z}_j$. Gram matrices are positive semi-definite, and scaling by the positive constant $\frac{1}{d}$ preserves this property.

To verify, for any coefficient vector $\boldsymbol{\alpha} = [\alpha_1, ..., \alpha_n]^T$:
\begin{equation}
    \boldsymbol{\alpha}^T \mathbf{K} \boldsymbol{\alpha} = \frac{1}{d} \boldsymbol{\alpha}^T \mathbf{Z} \mathbf{Z}^T \boldsymbol{\alpha} = \frac{1}{d} ||\mathbf{Z}^T \boldsymbol{\alpha}||^2 \geq 0.
\end{equation}

\emph{Proof of Symmetry:} The inner product operation is commutative:
\begin{align}
    K(x_i, x_j) &= <\mathbf{z}_i, \mathbf{z}_j> \nonumber \\ &= <\mathbf{z}_j, \mathbf{z}_i> \nonumber \\ &= K(x_j, x_i).
\end{align}

Having established both properties, the proximity kernel satisfies Mercer's conditions and defines a valid reproducing kernel Hilbert space.

\emph{Implications for Kernel Methods:} The validity of the proximity kernel enables its use in kernel-based algorithms. When representations $\mathbf{z}_i$ serve as features in k-means clustering, the algorithm minimizes:
\begin{equation}
    \arg\min_{\text{clusters}} \sum_{i} ||\mathbf{z}_i - \boldsymbol{\mu}_c||^2,    
\end{equation}

\noindent where $\boldsymbol{\mu}_c$ is the cluster centroid. Expanding the squared distance shows that this objective depends on kernel values $K(x_i, x_j)$, making k-means in representation space equivalent to a kernel-based clustering approach.

\subsection{Summary}

The proposed proximity kernel can be summarized as follows:
\begin{enumerate}
    \item Bin Center Selection: For each feature $j$, compute equal-frequency-based bin centers using observed values only.
    \item Proximity Assignment: Assign each observed value to its nearest bin center.
    \item One-Hot Encoding: Create binary encodings for observed features.
    \item Missing Value Handling: Apply the cascading fallback strategy using KME at each level.
    % \item Similarity Measure: Calculate pairwise similarities as the expected probability of bin co-occurrence across features.
\end{enumerate}

\section{Experiments}

\subsection{Experimental Settings}
\textbf{Datasets:} We evaluate the proximity kernel on 12 real-world datasets with varying characteristics. The datasets span multiple domains including medical diagnosis (Kidney, Hepatitis, Heart, Tumor, Cancer), biological classification (Soybean, Mushroom), financial analysis (Credit), demographic prediction (Adult), political analysis (Voting), and others (Mammography, Horse). These datasets exhibit diverse properties in terms of sample size and feature dimensionality, providing a comprehensive testbed for evaluating incomplete data methods. Table~\ref{tab:datasets} shows the statistical information of datasets. All of these datatsets are from UCI datasets~\footnote{https://archive.ics.uci.edu/datasets}.

\begin{table}[!htbp]
\caption{Statistical information of datasets}
\label{tab:datasets}
\resizebox{0.48\textwidth}{!}{%
\begin{tabular}{l|llllll}
\hline
Dataset     & \# Samples & \# Dim. & \# Cat. & \# Num.  & \# Missing & Class   \\ \hline
Kidney      & 400        & 25      & 3        & 22         & 10.54  & 2\\
Mammo       & 961        & 5       & 0       & 4        & 3.37      & 2\\
Heart       & 303        & 13      & 7       & 6        & 0.15      & 5\\ 
Hepatitis   & 155        & 19      & 13       & 6        & 5.67      & 2\\
Horse       & 368        & 27      & 0       & 27        & 19.39      & 2\\
Cancer      & 699        & 9      & 1       & 8        & 0.25      & 2\\
Credit      & 690        & 15      & 9       & 6        & 0.64      & 2\\ 
Adult       & 48842      & 14      & 7       & 7        & 0.32      & 4\\ 
Voting      & 435        & 16      & 16       & 0        & 5.63      & 2\\ 
Mushroom    & 8124       & 22      & 21       & 1        & 1.39      & 2\\ 
Tumor       & 339        & 17      & 0       & 17        & 3.90      & 21\\ 
Soybean     & 307        & 35      & 35       & 0        & 6.63      & 19\\ 
 \hline
\end{tabular}%
}
\end{table}

\textbf{Baselines:} We compare 9 representative methods : (1) Simple imputation: Mean imputation; (2) Statistical methods: MissForest, MICE, KNN, EM; (3) Deep learning approaches: GAIN, MIRACLE, MIWAE; (4) Kernel-based method: HI-PMK. These baselines cover the spectrum from classical to state-of-the-art techniques for handling incomplete data. The code of them are from python library scikit-learn~\footnote{https://scikit-learn.org/stable/modules/impute.html} and hyperimpute~\footnote{https://hyperimpute.readthedocs.io/en/latest/}~\cite{Jarrett2022HyperImpute}.

\textbf{Evaluation Protocol:} Following standard practice, we evaluate all methods on clustering quality using Normalized Mutual Information (NMI) as the metric. For fair comparison, we applied K-means with the number of clusters to the ground-truth number of classes. Each experiment was conducted 10 times and report average results for stability.

\textbf{Hyperparameter Settings}: To ensure fair comparison, we perform hyperparameter tuning for all baseline methods. MissForest is configured with mean as initial guess and maximum iterations searched over $\{5, 10, 15\}$. MICE explores maximum iterations in $\{10, 100, 200, 500\}$. KNN imputation searches the number of neighbors in $\{5, 10, 20\}$. The EM algorithm tunes maximum iterations over $\{10, 100, 200, 500\}$. For deep learning methods, GAIN searches training epochs in $\{100, 200, 500\}$, MIRACLE tunes learning rate over $\{0.001, 0.0001, 0.00001\}$, MIWAE explores epochs in $\{100, 200, 500\}$, and they use the same batch size in $\{32, 64, 128, 256\}$. For the proximity kernel, we search the number of bins per dimension in $\{2, 3, 4, 6, 8\}$.

\subsection{Empirical Evaluation}

\begin{table*}[!htbp]
\centering
\caption{Results between baselines and proposed methods in terms of NMI. The best performance are in bold.}
\label{result}
\resizebox{0.9\textwidth}{!}{%
\begin{tabular}{l|llllllllllll|l}
\toprule
Data       & Kidney          & Mammo           & Heart           & Hepatitis       & Horse           & Wiscom         & Credit          & Adult           & Voting & M.room       & Tumor  & Soybean   &AVG.      \\
\midrule
Mean       & 0.5457          & 0.2584          & 0.1660          & 0.1932          & \underline{0.2350}           & 0.7315         & 0.2101          & 0.0477          & 0.4935 & 0.1114         & 0.3679 & 0.7008   & 0.3384       \\
MissForest & 0.5597          & 0.264           & 0.1699          & 0.1978          & 0.1900          & 0.7427         & 0.2612          & 0.0605          & \underline{0.4958} & 0.0905         & 0.3728 & 0.7040      &0.3424    \\
MICE       & 0.5473          & 0.2624          & 0.1724          & 0.2082          & 0.1358          & 0.7427         & 0.2756          & 0.0506          & 0.4773 & 0.0957         & 0.3700 & 0.7061     & 0.3370     \\
KNN        & \underline{0.5696}          & 0.2636          & 0.1651          & 0.1861          & \textbf{0.2586} & 0.7427         & 0.3095          & 0.0542          & 0.4903 & 0.0799         & 0.3726 & 0.7145       &0.3506   \\
EM         & 0.5405          & 0.2693          & 0.1652          & 0.1739          & 0.1860          & 0.7039         & 0.2717          & 0.0638          & 0.4852 & 0.1213         & 0.3687 & 0.6954      &    0.3371\\
GAIN       & 0.5593          & 0.2723          & 0.1651          & 0.1989          & 0.1600          & 0.7427         & 0.2668          & 0.0729          & 0.4895 & 0.0556         & 0.3678 & 0.7134      &   0.3387 \\
MIRACLE    & 0.5385          & 0.2633          & 0.1985          & 0.1590          & 0.0147          & 0.7427         & 0.1582          & 0.0728          & 0.4840 & 0.0556         & 0.3653 & 0.7370       &0.3158   \\
MIWAE      & 0.5522          & 0.2897          & 0.1638          & 0.1590          & 0.2130          & 0.7295         & \textbf{0.4279} & 0.0680          & \textbf{0.5012} & 0.0559         & 0.3737 & 0.7137     & 0.3540     \\
HI-PMK     & \underline{0.6663}          & \underline{0.3271}          & \underline{0.2142}          & 0.2001          & 0.1006          & \underline{0.7585}         & 0.2316          & \textbf{0.0987} & 0.4892 & \underline{0.3061}         & \textbf{0.3872} & \underline{0.7478}     &\underline{0.3773}     \\
\midrule
Ours       & \textbf{0.6964} & \textbf{0.3315} & \textbf{0.2323} & \textbf{0.2289} & 0.1911          & \textbf{0.785} & \underline{0.3698}          & \underline{0.0791}          & 0.4947 & \textbf{0.554} & \underline{0.3774} & \textbf{0.7514}  & \textbf{0.4245}\\
\bottomrule
\end{tabular}%
}
\end{table*}

\begin{figure}[!htbp]
    \centering
    \includegraphics[width=0.95\linewidth]{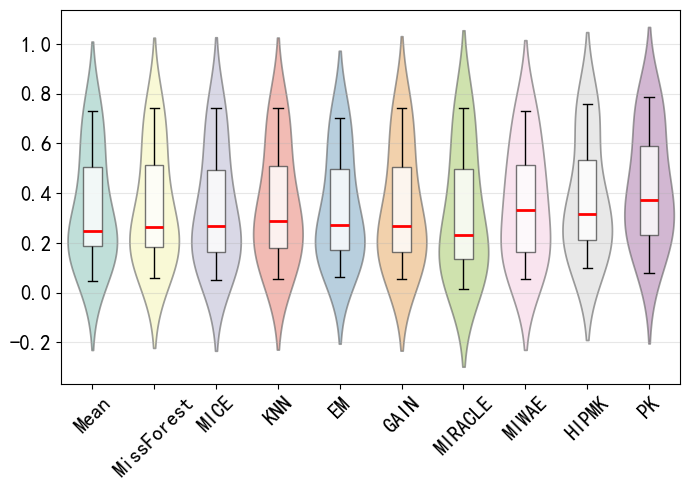}
    \caption{Comparison of all methods performance in terms of NMI.}
    \label{fig:boxplot}
\end{figure}

% Figure~\ref{boxplot} shows the performance comparison using combined box and violin plot visualizations. The boxplots reveal clear performance hierarchies: our proposed method achieves the highest median performance. The violin plots complement this analysis by revealing the underlying performance distributions. PK shows distributions concentrated in the upper performance range. In contrast, some other baseline methods exhibit wider spreads with substantial portions extending toward poor performance regions. This distribution analysis indicates that our method not only achieve higher average performance but also demonstrate more reliable performance across diverse dataset characteristics. 

Figure~\ref{fig:boxplot} presents the performance comparison using box and violin plots. The boxplots highlight a clear performance hierarchy, with the proposed method achieving the highest median. The violin plots reveal the shape of the performance distributions: PK is concentrated in the high-performance region, whereas several baseline methods display broader distributions, with significant mass extending into lower-performance ranges. This indicates that our method not only attains superior average performance but also maintains greater consistency across datasets with varying characteristics. 

Table~\ref{result} presents the full results across 12 datasets. The proposed method achieves the best or second-best performance on 10 out of 12 datasets, and obtains the highest average NMI across all datasets. The superior and stable performance of the proximity kernel highlights the effectiveness of computing similarity directly in the kernel representation space through the cascading fallback strategy, rather than relying on explicit imputation in the original feature space followed by separate similarity computation.

\begin{figure}[!htbp]
    \centering
    \includegraphics[width=0.95\linewidth]{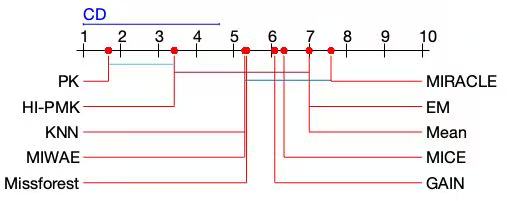}
    \caption{Friedman-Nemenyi test at significance level 0.1. If two algorithms are connected by a CD (critical difference) line, there is no significant difference between them.}
    \label{fig:nem}
\end{figure}

The critical difference diagrams in Figure~\ref{fig:nem} provide Friedman-Nemenyi test~\cite{demvsar2006statistical} at significance level $0.1$. PK is the only method that shows significant better performance than all methods except HI-PMK. This establishes the statistical superiority and robustness of the proposed approach across different datasets.

\subsection{Missing Rate Handling}

\begin{figure}[!htbp]
     \centering
     \begin{subfigure}[b]{0.95\linewidth}
         \centering
         \includegraphics[width=\textwidth]{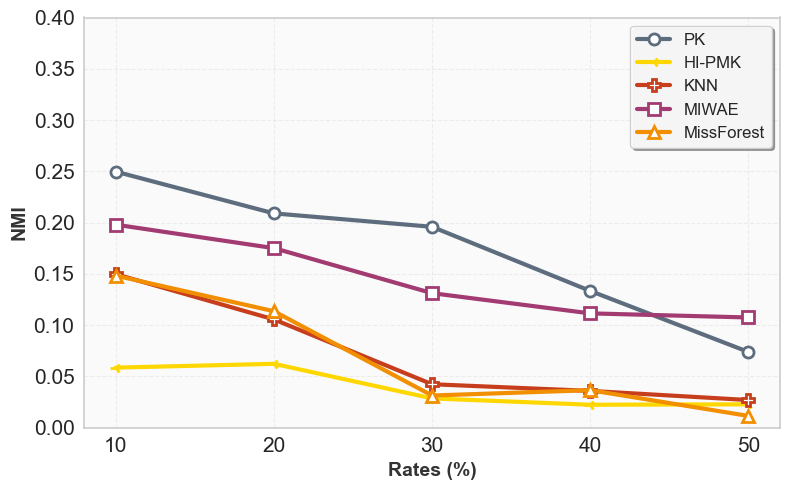}
         \caption{3L}
         \label{3L}
     \end{subfigure}
     
     \begin{subfigure}[b]{0.95\linewidth}
         \centering
         \includegraphics[width=\textwidth]{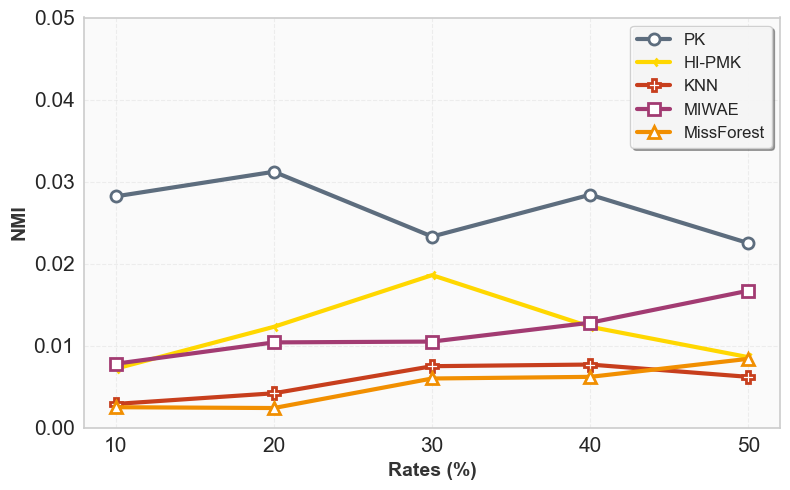}
         \caption{Messidor}
         \label{messi}
     \end{subfigure}
     \caption{Clustering performance of top 5 methods under different missing rates on 3L and Messidor datasets.}
        \label{rate}
\end{figure}

To evaluate the robustness of PK under varying degrees of data incompleteness, we conduct experiments on two non-missing datasets (3L and Messidor). We systematically generate missing data at different rates ranging from 10\% to 50\% using the Missing Completely At Random (MCAR) mechanism, where each value has an equal probability of being missing independent of both observed and unobserved data. This controlled experimental setup allows us to isolate the impact of missing rate on algorithm performance.

Figure~\ref{rate} presents the clustering performance measured by NMI across different missing rates for both datasets. On the 3L dataset (Figure~\ref{3L}), the proposed PK method maintains superior and relatively stable performance across low to moderate missing rates (10\%-40\%). Baseline methods such as KNN and MissForest exhibit steeper performance degradation, with their NMI scores dropping to near-zero by 30\% missing rate, indicating a complete loss of clustering quality. While most baseline methods struggle significantly as the missing rate increases, MIWAE shows competitive performance and slightly outperforms PK at the 50\% missing rate. The proposed method demonstrates greater resilience, sustaining reasonable performance even at higher missing rates. However, this marginal advantage comes at a significantly higher computational cost, MIWAE requires iterative neural network training with multiple epochs, while PK operates with linear time complexity through a single-pass algorithm. 

On the Messidor dataset (Figure~\ref{messi}), the proposed method gets best performance on all rates. At the extreme 50\% missing rate, all methods face substantial challenges due to severe information loss, with performance metrics declining across the board. Nevertheless, PK maintains a favorable efficiency-performance trade-off compared to computationally expensive deep learning approaches. The cascading fallback strategy proves effective in leveraging available information even when half of the data is missing, gradually relaxing from intersection-based matching to union-based matching and finally to global priors as the missing rate increases.

% These results demonstrate that the proximity kernel method not only achieves competitive accuracy but also provides practical scalability advantages for real-world web applications where computational efficiency is crucial alongside prediction quality.

\subsection{Scalability Analysis}

\begin{figure}[!htbp]
    \centering
    \includegraphics[width=0.95\linewidth]{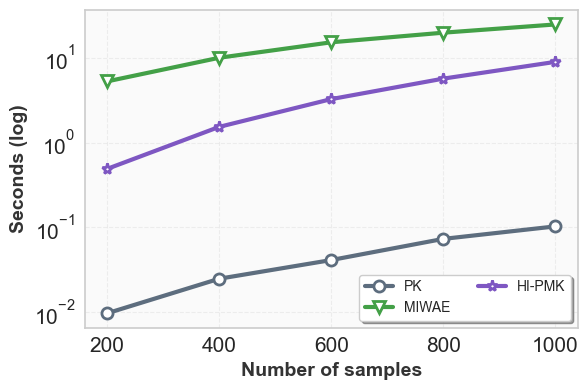}
    \caption{Runtime comparison between PK, MIWAE and HI-PMK.}
    \label{fig:time}
\end{figure}

We analyze the computational complexity of the proximity kernel in terms of both time and space requirements.

\textbf{Time Complexity}: PK consists of three main phases. The bin center selection phase computes percentiles for each feature, requiring $O(n \log n)$ time for sorting observed values per feature, yielding $O(d \cdot n \log n)$ overall. The encoding phase assigns each observed value to its nearest bin center in $O(n_{bins})$ time per value, resulting in $O(n \cdot d \cdot n_{bins})$ complexity. Since $n_{bins}$ is a small constant (typically 3-10), this is effectively $O(nd)$. The missing value handling phase has worst-case complexity $O(n^2 d)$ when union-based matching is frequently required. However, in practice, intersection-based matching succeeds for most samples, reducing the average complexity significantly. Overall, the time complexity is $O(nd + d \cdot n \log n)$, which is linear in the number of samples for fixed dimensionality. 

Figure~\ref{fig:time} presents a comparison of the running time for PK, HI-PMK, and MIWAE, where HI-PMK and MIWAE represent the best-performing traditional and deep learning methods, respectively. The results demonstrate that PK is significantly faster than both competing methods.

\textbf{Space Complexity}: The method requires $O(d \cdot n_{bins})$ space to store bin centers, $O(n \cdot d \cdot n_{bins})$ space for the encoded representations, and $O(n_{bins} \cdot d)$ space for global priors. Since $n_{bins}$ is a small constant, the space complexity is effectively $O(nd)$, linear in both the number of samples and features. The sparse nature of the representation (sparsity ratio $\frac{n_{bins}-1}{n_{bins}}$) enables efficient storage and computation.

\textbf{Practical Scalability}: The linear scaling in sample size makes the proximity kernel suitable for large-scale applications. Unlike methods requiring iterative optimization (EM, MICE) or expensive neural network training (GAIN, MIWAE), the proximity kernel computes representations in a single pass after the initial bin center selection. The representation construction is parallel across samples, allowing for straightforward distributed implementation for web-scale datasets.

\subsection{Sensitivity Analysis}

\begin{figure}[!htbp]
    \centering
    \includegraphics[width=0.95\linewidth]{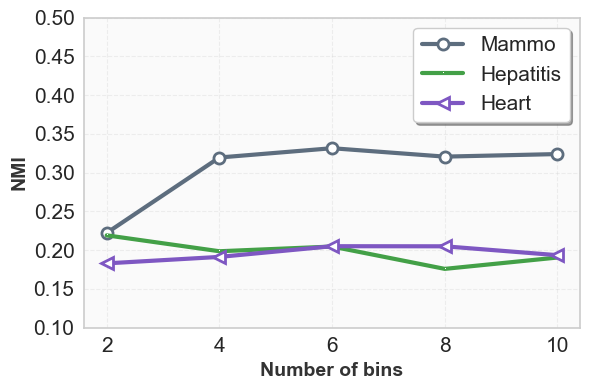}
    \caption{Performance under different number of bins.}
    \label{bins}
\end{figure}

The proximity kernel method has a single hyperparameter: the number of bins per feature ($n_{bins}$). Figure~\ref{bins} presents the clustering performance (NMI) across three datasets (Mammo, Hepatitis, and Heart) as the number of bins varies from 2 to 10.

The results demonstrate that the proximity kernel exhibits stable performance across different bin configurations. For the Mammo dataset, NMI scores fluctuate only slightly between 0.30 and 0.33 across all tested bin numbers. Similarly, the Heart and Hepatitis datasets maintain consistent performance around 0.18-0.2 and 0.17-0.21, respectively. This stability indicates that the proposed method is robust to hyperparameter choices and does not require extensive tuning to achieve good performance.

In practice, we found that $n_{bins}$ values between 3 and 6 provide a good balance between representation granularity and computational efficiency. Too few bins ($n_{bins}$ = 2) may result in overly coarse discretization, while too many bins may lead to sparse representations without meaningful performance gains.

This robustness contrasts with many baseline methods that require careful tuning of multiple hyperparameters, such as the number of neighbors in KNN, learning rates and epochs in deep learning approaches, or iterations in EM and MICE. The minimal parameterization makes the proximity kernel particularly suitable for large-scale applications where extensive hyperparameter search may be impractical.

\section{Discussion and Conclusion}

This paper presents proximity kernel, a novel similarity measure for incomplete data that integrates data dependent binning with a cascading fallback strategy. Unlike existing two-stage approaches that separate imputation from similarity computation, our method jointly handles missing values through kernel mean embedding over progressively relaxed matching criteria, from intersection-based to union-based to global priors. The equal-frequency binning combined with proximity assignment creates partitions that adapt to data density, while the resulting kernel representation is proven to be positive semi-definite and valid for kernel-based algorithms. Experimental results on 12 datasets demonstrate that the proximity kernel achieves superior performance with linear time complexity, making it suitable for large-scale web applications. 
% Future work could explore adaptive bin selection and theoretical analysis of the cascading fallback convergence properties.

It is worth noting that the current evaluation focuses on the Missing Completely At Random (MCAR) mechanism, where missingness is independent of both observed and unobserved values. Real-world data often exhibits more complex missing patterns, such as Missing At Random (MAR) and Missing Not At Random (MNAR). Investigating the proximity kernel's behavior under these more challenging scenarios represents an important direction for future work, along with exploring adaptive bin selection strategies and establishing theoretical convergence guarantees for the cascading fallback mechanism.

\newpage
% \subsection{Summary}

% The complete proximity kernel algorithm can be summarized as follows:

% 1. **Bin Center Construction**: For each feature $j$, compute quantile-based bin centers using observed values only.

% 2. **Proximity Assignment**: Assign each observed value to its nearest bin center.

% 3. **One-Hot Encoding**: Create binary encodings for observed features.

% 4. **Missing Value Handling**: Apply the cascading fallback strategy using kernel mean embedding principles at each level.

% 5. **Similarity Computation**: Calculate pairwise similarities as the expected probability of bin co-occurrence across features.

% The algorithm's effectiveness stems from the synergy between density-adaptive binning that respects data distribution, proximity assignment that creates meaningful neighborhoods, and the cascading fallback strategy that maximizes information utilization through kernel mean embedding principles.

% \begin{acks}
% \end{acks}

\bibliographystyle{ACM-Reference-Format}
\bibliography{main}

% \appendix



\end{document}